\documentclass[pdflatex,sn-mathphys-num]{sn-jnl}

\usepackage{graphicx}%
\usepackage{multirow}%
\usepackage{amsmath,amssymb,amsfonts}%
\usepackage{amsthm}%
\usepackage{mathrsfs}%
\usepackage[title]{appendix}%
\usepackage{xcolor}%
\usepackage{textcomp}%
\usepackage{manyfoot}%
\usepackage{booktabs}%
\usepackage{algorithm}%
\usepackage{algorithmicx}%
\usepackage{algpseudocode}%
\usepackage{listings}%

\usepackage{array} 

\newcommand{\authorbio}[3]{%
  \noindent
  \begin{tabular}[t]{@{}m{0.40\textwidth}m{0.70\textwidth}@{}}
    \includegraphics[width=\linewidth]{#2} &
    \textbf{#1}\ #3
  \end{tabular}\par\vspace{1.2em}%
}

\theoremstyle{thmstyleone}%
\theoremstyle{thmstyletwo}%

\theoremstyle{thmstylethree}%

\begin{document}

\title[Article Title]{ADHDeepNet From Raw EEG to Diagnosis: Improving ADHD Diagnosis through Temporal-Spatial Processing, Adaptive Attention Mechanisms, and Explainability in Raw EEG Signals}


\author[1]{\fnm{Ali} \sur{Amini}}\email{ali.amini@ktu.edu}

\author*[2]{\fnm{Mohammad} \sur{Alijanpour}}\email{alijanpour@ucf.edu}

\author[3]{\fnm{Behnam} \sur{Latifi}}\email{behnamlatifi72@gmail.com}

\author[4]{\fnm{Ali} \sur{Motie Nasrabadi}}\email{nasrabadi@shahed.ac.ir}

\affil[1]{\orgdiv{Centre of Real Time Computer Systems, Faculty of Informatics}, \orgname{Kaunas University of Technology}, \orgaddress{\street{K. Donelaičio g. 73}, \city{Kaunas}, \postcode{44249}, \country{Lithuania}}}

\affil*[2]{\orgdiv{Electrical Engineering and Computer Science Department}, \orgname{University of Central Florida}, \orgaddress{\street{Central Florida Blvd}, \city{Orlando}, \postcode{32816}, \state{Florida}, \country{USA}}}

\affil[3]{\orgdiv{Electrical Engineering Department}, \orgname{Amirkabir University of Technology}, \orgaddress{\street{Hafez}, \city{Tehran}, \postcode{15916-34311}, \state{Tehran}, \country{Iran}}}

\affil[4]{\orgdiv{Biomedical Engineering Department}, \orgname{Shahed University}, \orgaddress{\street{Persian Gulf}, \city{Tehran}, \postcode{33191-18651}, \state{Tehran}, \country{Iran}}}


\abstract{Attention Deficit Hyperactivity Disorder (ADHD) is a common brain disorder in children that can persist into adulthood, affecting social, academic, and career life. Early diagnosis is crucial for managing these impacts on patients and the healthcare system but is often labor-intensive and time-consuming. This paper presents a novel method to improve ADHD diagnosis precision and timeliness by leveraging Deep Learning (DL) approaches and electroencephalogram (EEG) signals. We introduce ADHDeepNet, a DL model that utilizes comprehensive temporal-spatial characterization, attention modules, and explainability techniques optimized for EEG signals. ADHDeepNet integrates feature extraction and refinement processes to enhance ADHD diagnosis. The model was trained and validated on a dataset of 121 participants (61 ADHD, 60 Healthy Controls), employing nested cross-validation for robust performance. The proposed two-stage methodology uses a 10-fold cross-subject validation strategy. Initially, each iteration optimizes the model's hyper-parameters with inner 2-fold cross-validation. Then, Additive Gaussian Noise (AGN) with various standard deviations and magnification levels is applied for data augmentation. ADHDeepNet achieved 100\% sensitivity and 99.17\% accuracy in classifying ADHD/HC subjects. To clarify model explainability and identify key brain regions and frequency bands for ADHD diagnosis, we analyzed the learned weights and activation patterns of the model's primary layers. Additionally, t-distributed Stochastic Neighbor Embedding (t-SNE) visualized high-dimensional data, aiding in interpreting the model's decisions. This study highlights the potential of DL and EEG in enhancing ADHD diagnosis accuracy and efficiency.}




\maketitle

\section{Introduction}\label{sec1}

ADHD ranks among the leading neurodevelopmental disorders, characterized by indications of inattentiveness, excessive activity, and impulsive behavior, impacting both children and enduring into adult life \cite{Talebi2022,DeDea2019}. The disorder is characterized by two distinct aspects represented by the two words in its name. The first symptom is attention deficit, where individuals struggle to maintain focus on a particular topic and are easily distracted by their surroundings. The second symptom is hyperactivity, where individuals have difficulty remaining still and often feel the need to constantly change their posture or position. This can manifest as sudden outbursts or moments of silence during regular conversations \cite{TaghiBeyglou2022}. Around 5\% of children globally are diagnosed with ADHD, and research indicates that the condition is more common in boys (12.1\%) compared to girls (3.9\%) \cite{Mohammadi2016,Ekhlasi2023}. In fact, as reported by the National Institute of Mental Health (NIMH), approximately 70\% of adolescents diagnosed with ADHD still display some degree of hyperactivity and impulsivity during their teen years and into adulthood \cite{TaghiBeyglou2022}.

There are several factors that cause ADHD. Genetics is one of the most important causes of this disorder. It is shown that a member of family diagnosed with ADHD can increase the risk of developing this condition in other members 5 to 10 times. There are also factors in the environment that can increase the chances of ADHD developing in a child, such as maternal smoking and alcohol consumption, low birth weight, preterm delivery, and contact with particular pollutants in the environment \cite{Tanko2022}. The early detection of ADHD plays a crucial role in effectively treating patients. Typically, the diagnosis of ADHD relies on behavioral assessments and conversations. Yet, research has shown that numerous general practitioners do not have adequate understanding to properly identify ADHD. Among the 400 primary care physicians surveyed, 44\% indicated that the criteria for diagnosing ADHD were ambiguous. Further, 72\% indicated that diagnosing ADHD in youths was simpler compared to in grown-ups, and 75\% deemed the reliability of the ADHD diagnostic standards to be either subpar or fair \cite{Moghaddari2020}. As these diagnosis methods are time-consuming and based on the person’s behavior during interview sessions, there is a need for developing more accurate and reliable methods for ADHD detection. 

It is shown that brain disorders are usually caused due to different brain functions which are as a result of differences in brain regions’ connectivity. Therefore, employing neuroimaging techniques to assess brain connectivity for diagnosing disorders can be more reliable because these methods address the pathophysiology of neuropsychiatric conditions from a systemic viewpoint \cite{Talebi2022,Ekhlasi2021}. Several neuroimaging modalities are utilized to assess brain connectivity, including functional magnetic resonance imaging (fMRI), electroencephalography (EEG), and magnetoencephalography (MEG) \cite{Riaz2020}. The authors in \cite{Deshpande2015} used a fully connected cascade (FCC) artificial neural network (ANN) framework on fMRI data. Their proposed network achieved 90\% and 95\% accuracy in differentiating ADHD from healthy individuals and among the ADHD subtypes, respectively. In \cite{Riaz2020}, a DL-based architecture called DeepFMRI is proposed for ADHD detection. The architecture includes a CNN as the feature extractor, Siamese-inspired similarity measure networks as the functional connectivity network, and a fully connected network as the classifier. Their architecture achieved classification with 73.1\% accuracy, 91.6\% specificity, and 65.5\% sensitivity on the ADHD-200 dataset from the New York University imaging site.

Despite very high spatial resolution of fMRI images, they are very expensive to achieve. The EEG modality can also offer high temporal resolution and excellent spatial resolution in high-density electrode recordings. Moreover, EEG recordings can be captured during physical activity and everyday routines, including those involving children exhibiting hyperactivity \cite{Bakhshayesh2019}. These advantages have motivated numerous investigators to employ EEG signals in their ADHD-focused investigations. Authors in \cite{Tor2021} proceeded to extract non-linear features from these coefficients to identify ADHD, conduct disorder (CD), and ADHD in conjunction with CD. They examined EEG data collected from 123 children (45 diagnosed with ADHD, 62 with conduct disorder + ADHD, and 16 with conduct disorder) using a K-Nearest Neighbor classifier, achieving an accuracy rate of 97.88\%. In \cite{Maniruzzaman2022}, the researchers utilized the t-test (p-value < 0.05) and the least absolute shrinkage and selection operator (LASSO) to pinpoint promising features within the EEGs of children affected `by ADHD, to enhance classification accuracy. The findings revealed that combining LASSO with the SVM classifier yielded the highest accuracy of 94.2\%, a sensitivity of 93.3\%, and an F1-score of 91.9\%. Additionally, they attained 93.4\% accuracy, 91.7\% sensitivity, and a 91.1\% F1-score when using the Multilayer Perceptron (MLP) classifier.

One of the limits of earlier research on ADHD detection using EEG signals, including those explained above, is that first, they need to extract features, be they linear or non-linear, from EEGs and then feed the extracted features to the classifier. DL approaches address this challenge due to their automatic feature extraction. In fact, by using DL-based methods, the need for the pre-processing phase is minimized and in some cases, eliminated. CNNs have been the most widely used DL method in the field of ADHD detection in recent years. The primary challenge in applying CNNs to EEG signals is transforming the spatiotemporal properties of EEG signals, captured via numerous electrode channels, into 3-channel image-like formats \cite{Ekhlasi2021}. In \cite{Dubreuil2020}, the spectrogram of the EEG signals from 40 participants (20 with ADHD and 20 HC) was generated during the pre-processing phase. The resulting images from the spectrogram of the EEG signals were then used as input for a custom CNN for classification. The developed model achieved an 88\% classification accuracy. In \cite{Chen2019}, the authors used EEG signals from 50 children with ADHD and 51 HC children, proposing a DL framework to detect ADHD in children by converting EEG signals into image data and feeding them into their CNN architecture. The developed model achieved a 94.67\% accuracy on the test dataset. A 13-layer CNN for ADHD detection in children is also proposed in \cite{Ekhlasi2021}. The authors utilized the EEG signals of 61 children (31 with ADHD and 30 without) and removed the noise from these signals. Then they converted the signals into RGB images suitable for feeding to the CNN network. They employed the 5-fold cross-validation method and achieved an average accuracy of 99.06\% from their model. The research by \cite{Ahmadi2021} presents a deep neural network architecture aimed at classifying two subsets of ADHD children (ADHD-I, ADHD-C) from HC children using their resting EEG signals. First, they decomposed the signals into recognized frequency bands and constructed a collection of optimal spatial filters for each frequency band. Their network showcased excellent performance utilizing just three higher-frequency bands ($\beta_1$, $\beta_2$, $\gamma$), with accuracy, recall, and precision of 99.46\%, 99.45\%, and 99.48\%, respectively.

However, use of DL in ADHD detection from EEG signals does not limit to CNNs. In \cite{Tosun2021}, the authors utilized power spectral densities and spectral entropy measures from the EEG signals collected from subjects both with and without ADHD. Subsequently, they applied long short-term memory (LSTM), support vector machine (SVM), and ANN classifiers to the data, with the LSTM achieving the highest accuracy of 88.88\% on the "Fp1, F7" channel and 92.15\% during the eyes-closed resting condition. Their findings indicated that spectral entropy positively influences accuracy. Other ADHD-focused works have highlighted the value of PSD-derived brain maps with Siamese CNNs and Grad-CAM interpretability \cite{Latifi2024}. Beyond ADHD, temporal–spatial convolutional residual networks have also been successfully applied to EEG decoding in other neurological conditions \cite{Mirzabagherian2023}.

In this study, we introduce a unique approach by integrating nested cross-validation for hyper-parameter optimization with data augmentation (DA) to counter overfitting and improve model generalization. Additionally, we incorporate a model interpretability method to ensure decisions adhere to neuroscience and clinical principles. This combination of techniques, rare in existing studies, yields a robust, reliable, and interpretable model, significantly advancing the field.
The structure of this paper is delineated as follows: Section II elucidates the process of data collection and pre-processing, the architecture of the proposed model, and a comprehensive definition of the employed methodology. Section III presents the results derived from the implementation of the proposed model. Section IV engages in a comprehensive discussion of the proposed method and its corresponding results. Finally, Section V summarizes the conclusions derived from the study and highlights potential directions for future research.



\section{Methods and Materials}\label{sec2}
\subsection{Data Collection and Preprocessing}\label{subsec2}

The dataset used in this work is collected from a total number of 121 individuals, 60 of whom are HC and 61 are ADHD children \cite{Nasrabadi2018}. The participants in our research were boys and girls aged between 7 and 12 years, and those identified as having ADHD were diagnosed using the DSM-IV criteria by a seasoned psychiatrist. HC group children have not experienced any hard head injuries, epilepsy, drug abuse, psychiatric disorder and have never had any high-risk behavior. Given that ADHD influences visual focus, we employed a method involving displaying 
figures to the children and asking them to enumerate them, as depicted in Fig. 1. The EEG recordings were conducted at the Psychology and Psychiatry Research Center located within Roozbeh Hospital in Tehran, Iran. The EEG data can be described as multivariate time series. Specifically, our dataset includes a total of C=19 channels, which correspond to the number of electrodes used during the recording process: Fz, Cz, Pz, C3, T3, C4, T4, Fp1, Fp2, F3, F4, F7, F8, P3, P4, T5, T6, O1, O2, with a sampling rate of 128 Hz. It's worth noting that the amount of EEG data gathered from each participant may vary slightly. To represent the amplitude values measured from multiple EEG channels over time for individual subjects, we introduce a notation that captures the essence of the multivariate time series data inherent in EEG recordings. We define:

\begin{equation}
X_{(j)}^{c,i} =
\begin{bmatrix}
x_{(j)}^{1,1} & x_{(j)}^{2,1} & \cdots & x_{(j)}^{p_j,1} \\
x_{(j)}^{1,2} & x_{(j)}^{2,2} & \cdots & x_{(j)}^{p_j,2} \\
\vdots & \vdots & \ddots & \vdots \\
x_{(j)}^{1,C} & x_{(j)}^{2,C} & \cdots & x_{(j)}^{p_j,C}
\end{bmatrix}
\tag{1}
\end{equation}

\begin{figure}[t]
\centering
\includegraphics[width=0.5\textwidth]{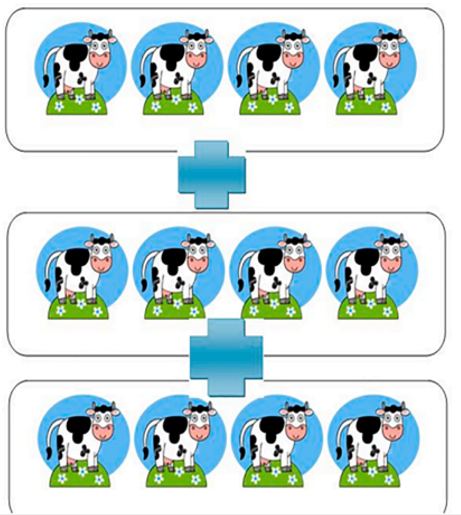}
\caption{An example of pictures shown to participants}\label{fig1}
\end{figure}

Here, $X$ stands for the EEG data matrix, $c$ indicates the channel index (of which there are $C$ channels in total), $i$ denotes the time index within the subject's EEG recording, and $j$ represents the subject index. Thus, the complete EEG data matrix for the $j$th subject is denoted by $X_{(j)}^{c,i} \in \mathbb{R}^{C \times p_j}$. In the above notation, $p_j$ signifies the total number of time indices in the data matrix for the $j$th subject, which corresponds to the length of EEG recording for that subject.

To accommodate DL approaches which necessitate ample training data, we segment each subject's EEG signals into four-second epochs. Given a sampling frequency $f_s = 128~\mathrm{Hz}$, a segment, or epoch, consists of $T = 4 \times 128 = 512$ time indices.

For segmentation of the subject $j$'s EEG data, we employ the following notation to express the $j$th segment:

\begin{equation}
X_{(j)}^{s} =
\begin{bmatrix}
x_{(j)}^{(s-1)T+1,1} & x_{(j)}^{(s-1)T+2,1} & \cdots & x_{(j)}^{sT,1} \\
x_{(j)}^{(s-1)T+1,2} & x_{(j)}^{(s-1)T+2,2} & \cdots & x_{(j)}^{sT,2} \\
\vdots & \vdots & \ddots & \vdots \\
x_{(j)}^{(s-1)T+1,C} & x_{(j)}^{(s-1)T+2,C} & \cdots & x_{(j)}^{sT,C}
\end{bmatrix}
\tag{2}
\end{equation}

The EEG data for the $s$th segment of each subject is denoted as $X_{(j)}^{s}$, where $s$ ranges from $1$ to $S_j$, and $S_j$ represents the overall number of segments for the $j$th individual. We operate under the assumption that we receive a single EEG dataset for each subject $j$. These datasets are divided into labeled trials, which are segments of the original recording categorized into one of two groups (ADHD or HC).

Concretely, we are given datasets $D^j = \{ (X_{(j)}^1, y_{(j)}^1),\, \ldots,\, (X_{(j)}^{S_j}, y_{(j)}^{S_j}) \}$, where $S_j$ signifies the total count of trials recorded for subject $j$. The input matrix $X_{(j)}^s \in \mathbb{R}^{E \times T}$ for trial $s$ encompasses signals from $E$ electrodes and $T$ discrete time intervals per trial. The class label associated with trial $s$ is indicated by $y_{(j)}^s$, taking its value from the set $\mathcal{L}$ comprising $K$ different class labels. Given our classification scenario, which distinguishes between ADHD and HC, $\forall\, y_{(j)}^s:\, y_{(j)}^s \in \mathcal{L} = \{ l_1 = \text{``ADHD''},\, l_2 = \text{``HC''} \}$.

The ground truth label of each subject is represented by a binary label vector, $Y_{(j)}$, where $Y_{(j)} = (1,0)$ corresponds to an ADHD subject and $Y_{(j)} = (0,1)$ denotes an HC subject. The label vector for each segment $X_{(j)}^s$ is given by
\begin{equation}
Y_{(j)}^s =
\begin{cases}
(1,0) & \text{if } y_{(j)}^s = \text{``ADHD''} \\
(0,1) & \text{if } y_{(j)}^s = \text{``HC''}
\end{cases}
\tag{3}
\end{equation}

For each segment, $X_{(j)}^s$, a classifier predicts the label vector $\hat{Y}_{(j)}^s$, which has the same dimensions as $Y_{(j)}$ and represents the predicted probabilities for each class. The ultimate predicted label for each subject is determined by aggregating the predicted labels of all segments linked to that subject, with the class exhibiting the highest cumulative probability being chosen. Mathematically, this is expressed as
\begin{equation}
\hat{Y}_{(j)} = \arg\max \left( \sum_{s=1}^{S_j} \hat{Y}_{(j)}^s \right)
\tag{4}
\end{equation}
where $\hat{Y}_{(j)}$ is the final predicted label for the $j$th subject.

The aim of this study is to leverage the gathered data to develop a predictive model that categorizes new instances belonging to either the ADHD or HC groups. For the purpose of this study, let us define the dataset $D = D^1 \cup D^2 \cup \ldots \cup D^{121}$. Within this combined dataset, the ADHD group is represented by 2330 samples, while the HC group accounts for 1843 samples. Furthermore, we did not apply any additional preprocessing steps to the samples $X_{(j)}^i \in D$. It is crucial to emphasize that the dataset employed in this work was constructed without the utilization of overlapping samples. Using overlapped EEG signals has been shown to give rise to overfitting as a consequence of the increased correlation between samples, accentuated focus on specific features, and the low diversity nature of data. Consequently, to ensure the development of a robust and generalizable model, the present study opted not to employ overlapped data.

\subsection{Model Explanation}\label{subsec2}

In this paper, we introduce an enhanced version of the EEGNet \cite{Lawhern2018} model, named ADHDeepNet, which provides significant benefits in the classification of ADHD. The EEGNet model has been shown to be proficient in capturing temporal dynamics and spatial information while optimizing feature extraction. One of the key benefits of EEGNet is its ability to operate directly on raw EEG signals, which eliminates the need for extensive preprocessing steps, thereby expediting the workflow and enhancing analytical efficiency. In this work, we introduce tailored modifications to the EEGNet architecture to further enhance its performance for ADHD classification. These modifications effectively align the architecture with the specific demands of ADHD classification, resulting in improved accuracy and robustness in detecting ADHD-related patterns within EEG signals. To quantify these enhancements, ADHDeepNet is characterized by a total of 228,642 parameters.

EEG signals are not usually fed to the network in a direct manner. As seen in the literature, majority of research use various features extracted from EEG signals as inputs of their models. As in EEGNet, there is also no feature extraction phase in this research since we aim to use raw EEG signals as the input for our proposed model. Eliminating feature extraction have two major impacts on the whole procedure. First, it speeds up the total process both in train and test phase, and second, it makes the system less dependent on human decisions (since the type of features to extract is determined by human expert), thus making the system more autonomous.

In designing ADHDeepNet, we were inspired by EEGNet as well as by multi-branch architectures such as Inception \cite{Szegedy2016}, Xception \cite{Chollet2017}, and SE blocks \cite{Hu2018}. Multi-branch modeling has previously been shown effective in domains outside EEG, particularly in video recognition, where two-stream CNNs demonstrated that complementary temporal and spatial representations can be learned in parallel \cite{Alijanpour2021}. Adopting a similar principle allows ADHDeepNet to better capture the joint spatio-temporal dependencies in EEG data.

\subsubsection{Temporal Dynamics and Spatial Information Extraction}\label{subsubsec2}

As illustrated in Fig. 2(a), the initial stage entails extracting the temporal features of the signal by employing a temporal 2-D convolutional (Conv) layer. This layer is applied to the temporal segments of the signal. Subsequently, a batch normalization (BN) layer is utilized to expedite model convergence, followed by a depthwise Conv layer.  Depthwise Convolution is a type of layer where each input channel is assigned its distinct kernel.

The spatial features of the EEG signals are isolated by applying this kernel individually to each of the 19 input channels. The BN layer alongside the Exponential Linear Unit (ELU) are implemented after the preceding layer. Next, an Average Pooling (AP) layer is deployed, which diminishes the output volume by fifty percent because the layer parameters are configured to (1×2). Finally, a dropout layer is incorporated to safeguard the proposed model against overfitting.

\begin{figure}[t]
\centering
\includegraphics[width=1\textwidth]{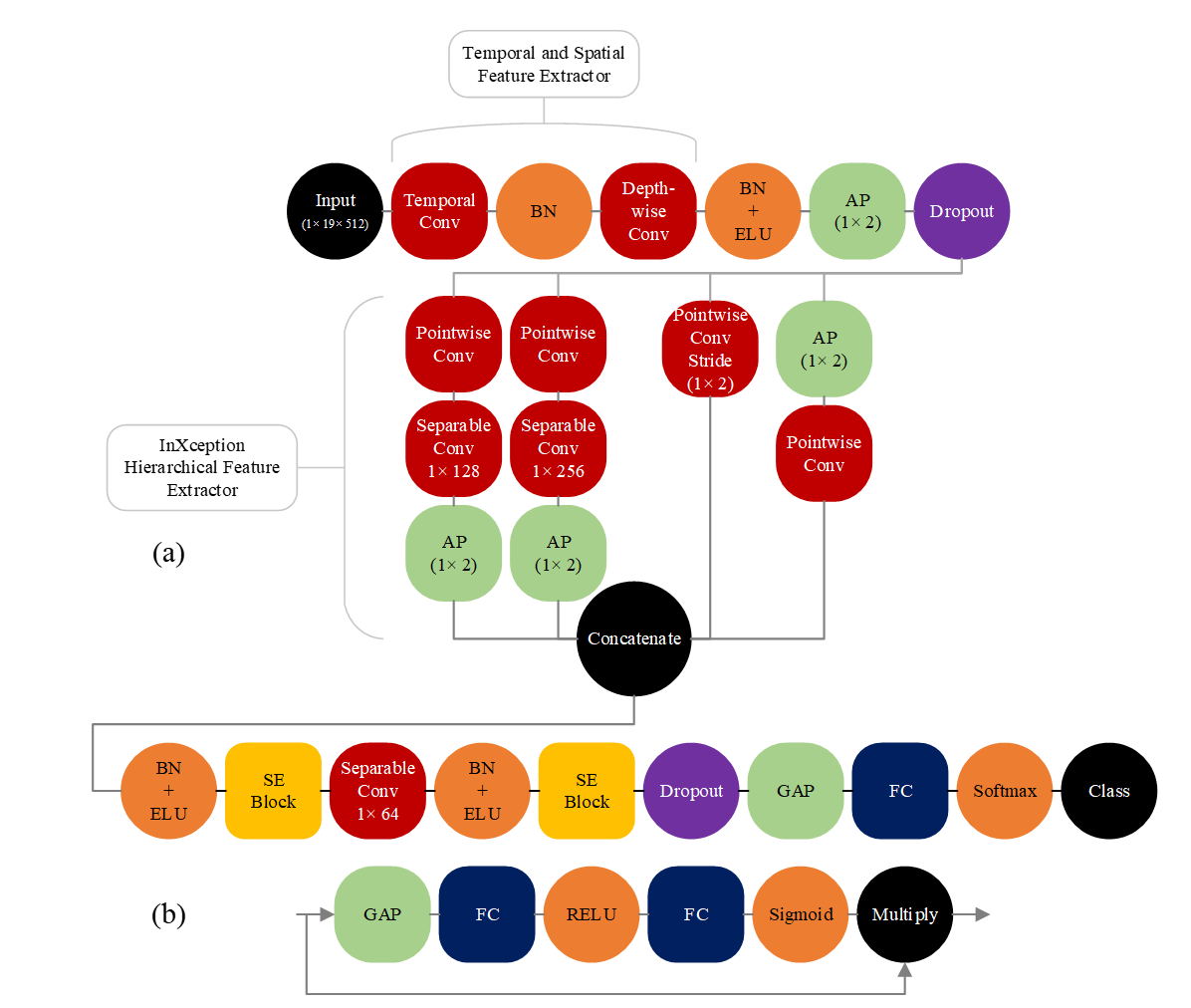}
\caption{(a): ADHDeepNet, (b): SE block (Channel-wise Feature Recalibrator)}\label{fig2}
\end{figure}

\begin{figure}[h]
\centering
\includegraphics[width=1\textwidth]{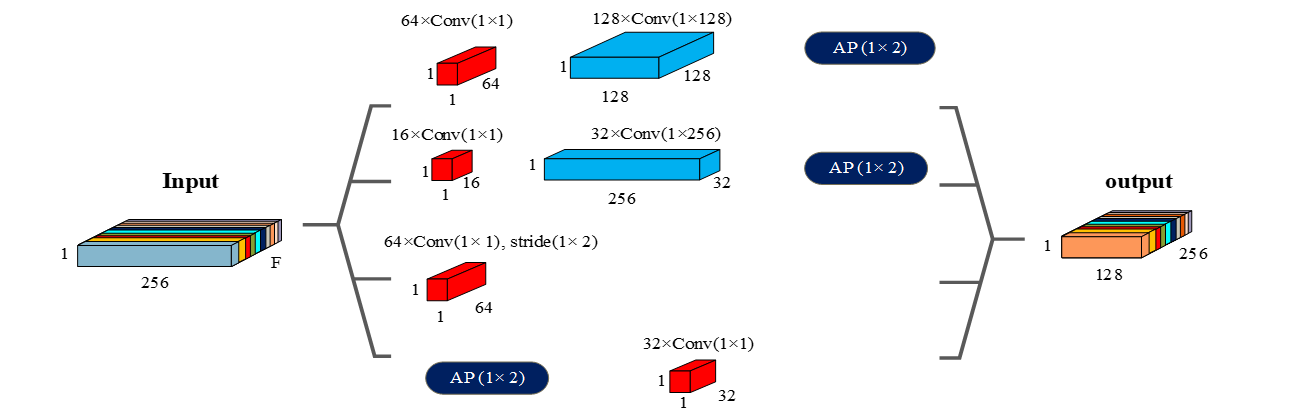}
\caption{The InXception module of ADHDeepNet}\label{fig3}
\end{figure}

\subsubsection{Hierarchical Features with Inxception}\label{subsubsec2}

To improve the proposed model, inspiration is drawn from the Inception and Xception models (InXception), which enables learning features at different scales and spatial resolutions, capturing a richer set of features in the input data. A comprehensive approach is employed for feature extraction, involving four parallel streams, which are combined through concatenation.  In two of these pathways, attributes obtained from earlier layers are combined through point-wise Conv kernels, succeeded by 2-D separable Conv layers featuring filter dimensions of 1×128 and 1×256.  First presented in the Xception architecture, these components efficiently decrease the quantity of model variables in comparison to traditional Conv layers.

\begin{figure}[h]
\centering
\includegraphics[width=1\textwidth]{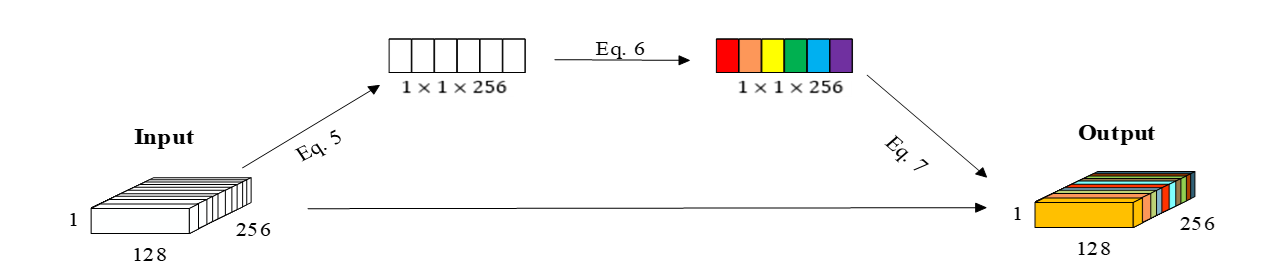}
\caption{The SE block of ADHDeepNet}\label{fig4}
\end{figure}

\subsubsection{Channel-Wise Feature Recalibration with SE Block}\label{subsubsec2}

We leveraged the principle of the SE block framework to enhance the classification precision within our suggested model. Fig. 2(b) illustrates the configuration of the SE blocks present in the model. This component is segmented into three parts: squeeze, excitation, and re-calibration, each of which we will detail further below.

\textbf{Squeeze operation:} The objective of the squeeze operation is to gather comprehensive spatial details through consolidating feature maps over their spatial extents ($H \times W$). Accomplishment of this is facilitated by employing a Global Average Pooling (GAP) layer. Consider $X \in \mathbb{R}^{C \times H \times W}$ as the input feature maps, where $C$ denotes the count of channels, $H$ represents the vertical dimension, and $W$ signifies the horizontal extent. The squeeze operation is shown in (5):

\begin{equation}
s_c = \frac{1}{H \cdot W} \sum_h \sum_w X_c(h,w), \quad \forall c \in \{1, \ldots, C\}
\tag{5}
\end{equation}

Where $s_c$ is the $c^{th}$ element of the channel-wise descriptor $s \in \mathbb{R}^C$, and $X_c(h,w)$ is the value at spatial location $(h,w)$ in the $c$th channel of the input feature maps.

\textbf{Excitation operation:} The excitation operation aims to learn non-linear relationships between channels and generate channel-wise weights. This process is realized through a two-layer fully connected (FC) network equipped with a ReLU in the intermediate layer and a sigmoid in the final layer. The excitation function can be defined as:

\begin{equation}
F_{ex}(s) = \sigma \big( W_2 \cdot \delta(W_1 \cdot s) \big)
\tag{6}
\end{equation}

where $\sigma$ represents the sigmoid, $\delta$ symbolizes the ReLU, $W_1 \in \mathbb{R}^{\frac{C}{r} \times C}$ and $W_2 \in \mathbb{R}^{C \times \frac{C}{r}}$ are the weights of the FC layers, and $r$ is the reduction ratio (a hyperparameter that controls the capacity of the SE block).

\textbf{Re-weighting operation:} The re-weighting operation applies the learned channel-wise weights to the input feature maps. This can be defined as:

\begin{equation}
Y_c = F_{ex}(s_c) \cdot X_c, \quad \forall c \in \{1, \ldots, C\}
\tag{7}
\end{equation}

where $Y_c$ is the $c^{th}$ channel of the output feature maps $Y \in \mathbb{R}^{C \times H \times W}$. Such attention-like mechanisms have also proven useful in multimodal physiological frameworks, such as MultiHeart, where robustness to missing or noisy inputs was achieved through recalibrating feature salience \cite{Ahmadi2025}. Incorporating the SE block into the model similarly enabled it to adaptively recalibrate the channel-wise feature responses based on the learned importance of each channel.This can help the model focus on the most relevant features in the EEG data.

\subsubsection{Enhanced Model Design and Overfitting Prevention}\label{subsubsec2}
Building upon the initial implementation of the SE block, the suggested approach maintains the application of a separable 2-D Conv layer measuring (1×64), succeeded by a BN layer and an ELU. The SE block is reapplied for the purpose of recalibrating the extracted feature maps.

To prevent overfitting, a dropout layer is incorporated into the architecture. Furthermore, instead of incorporating a flattened layer, which might result in an expansion of the model parameters, a GAP layer is adopted to promote a streamlined model architecture.

The final stage of the proposed method is using an FC layer with number of nodes proportional to the dataset classes. Softmax is employed in this FC layer to facilitate the classification of ADHD or HC groups.

\subsection{Problem Definition}\label{subsec2}

In a formal sense, under our supervised classification framework, the proposed model calculates a mapping from input data to a single real value for each class, 
$f(X^i; \theta): \mathbb{R}^{E \cdot T} \rightarrow \mathbb{R}^K$, 
with $\theta$ being the function's parameters, $E$ indicating the electrode count, $T$ signifying the duration of time intervals, $N$ representing the number of EEG trials, and $K$ denoting the potential outcome categories. To apply the model in classification tasks, the output is commonly converted into conditional likelihoods for a label $l_k$ based on the input $X^i$ through the softmax function:

\begin{equation}
p(l_k \mid f(X^i; \theta)) = \frac{\exp\!\left(f_k(X^i; \theta)\right)}{\sum_{k=1}^K \exp\!\left(f_k(X^i; \theta)\right)}
\tag{8}
\end{equation}

For educating the model to allocate higher probabilities to accurate labels by reducing the aggregate of individual sample losses:

\begin{equation}
\theta^* = \arg\min_{\theta} \sum_{i=1}^N \sum_{k=1}^2 -\log \big(p(l_k \mid f_k(X^i; \theta))\big) \cdot y^i
\tag{9}
\end{equation}

Based on the limited data available, we assessed the model through a (10-2)-fold cross-subject validation technique. The dataset $D$ was split into 10 non-overlapping folds $(D_1, D_2, \ldots, D_{10})$. This ensures that all trials from a particular individual are contained within a single fold. Additionally, the proportion of trials between the ADHD and HC remains nearly equal across the various folds and the entire dataset.

During the (10-2)-fold cross-subject validation procedure, one segment is marked as the evaluation set, termed $D_{cv}$, whereas the leftover dataset serves for training purposes, labeled $D_{train}$. The training data for each step of 10-fold cross-subject validation is divided into two non-overlapping parts called $D_{train}^1$ and $D_{train}^2$ with sample sizes of $N_1$ and $N_2$, respectively. By employing the 2-fold cross-subject validation strategy algorithm shown in Algorithm~1, and the Bayesian Optimization (BO) algorithm shown in Algorithm~2, the model's hyper-parameters are tuned through a specified number of iterations, preparing it for the testing phase which is elaborated in Algorithm~3. 

\begin{algorithm}[t]
\caption{Training Procedure Without Data Augmentation}
\label{alg:bo-nestedcv}
\begin{algorithmic}[1]
\Require Dataset $D$, CV folds $\{D_{cv}\}_{cv=1}^{10}$, BO iterations $T=100$
\Ensure Best hyperparameter configuration $\theta_h^{\star}$

\For{$cv = 1,2,\dots,10$} 
  \State $D_{\text{train}} \gets D \setminus D_{cv}$
  \For{$t = 1,2,\dots,100$}
    \Statex \textbf{(Split inner folds)}
    \State Choose a bipartition of $D_{\text{train}}$ into $D_{\text{train}}^{1}$ and $D_{\text{train}}^{2}$ such that
    \Statex \hspace{1.7em} $D_{\text{train}}^{1} \cup D_{\text{train}}^{2} = D_{\text{train}}, \quad D_{\text{train}}^{1} \cap D_{\text{train}}^{2} = \emptyset$
    \Statex

    \Statex \textbf{(Inner fold 1: train on $D_{\text{train}}^{1}$, validate on $D_{\text{train}}^{2}$)}
    \State $\displaystyle \theta_{1}^{\ast\,t} \gets \arg\min_{\theta_1^{t}} 
      \sum_{i=1}^{N_1}\sum_{k=1}^{2} \bigl(-\log p\bigl(\ell_k \mid f_k(X_1^{i}; \theta_h^{t}, \theta_1^{t})\bigr)\bigr)\,y_1^{i}$
    \Statex \hspace{1.7em} where $(X_1^{i}, y_1^{i}) \in D_{\text{train}}^{1}$, $i=1,\dots,N_1$
    \Statex

    \Statex \textbf{(Inner fold 2: train on $D_{\text{train}}^{2}$, validate on $D_{\text{train}}^{1}$)}
    \State $\displaystyle \theta_{2}^{\ast\,t} \gets \arg\min_{\theta_2^{t}} 
      \sum_{i=1}^{N_2}\sum_{k=1}^{2} \bigl(-\log p\bigl(\ell_k \mid f_k(X_2^{i}; \theta_h^{t}, \theta_2^{t})\bigr)\bigr)\,y_2^{i}$
    \Statex \hspace{1.7em} where $(X_2^{i}, y_2^{i}) \in D_{\text{train}}^{2}$, $i=1,\dots,N_2$
    \Statex

    \Statex \textbf{(Inner-fold accuracy)}
    \State $\displaystyle \text{Accuracy}(\theta_h^{t},\theta_{1}^{\ast\,t},\theta_{2}^{\ast\,t},X_1,y_1,X_2,y_2)
      = \frac{1}{N_1+N_2}\Bigg[
        \sum_{i=1}^{N_1} \delta\!\Big(\arg\max_{k} p\!\big(\ell_k \mid f_k(X_1^{i};\theta_h^{t},\theta_{1}^{\ast\,t})\big)=y_1^{i}\Big)
        + \sum_{i=1}^{N_2} \delta\!\Big(\arg\max_{k} p\!\big(\ell_k \mid f_k(X_2^{i};\theta_h^{t},\theta_{2}^{\ast\,t})\big)=y_2^{i}\Big)
      \Bigg]$
    \Statex \hspace{1.7em} ($\delta$ is the Kronecker delta; returns $1$ if the predicate is true, else $0$)
    \Statex

    \Statex \textbf{(BO objective and hyperparameter update)}
    \State $\displaystyle g(\theta_h^{t}) \gets -\,\text{Accuracy}(\theta_h^{t},\theta_{1}^{\ast\,t},\theta_{2}^{\ast\,t},X_1,y_1,X_2,y_2)$
    \State Select next $\theta_h^{t+1}$ using Bayesian Optimization guided by $g$
  \EndFor
\EndFor
\State \textbf{return} $\theta_h^{\star}$ (best over BO/CV according to validation performance)
\end{algorithmic}
\end{algorithm}

Let $S$ denote the search space, which is a Cartesian product of the individual hyper-parameter domains:

\begin{equation}
S = d_1 \times d_2 \times \ldots \times d_n
\tag{10}
\end{equation}

Here, $n$ represents the number of hyper-parameters, and $d_i$ is the domain of the $i$th hyper-parameter. For our specific problem, the search space $S$ includes the domain of hyper-parameters. The hyper-parameters of the proposed model, denoted by $\theta_h \in S$, include the norm-rate, the learning rate, the optimizer type, the dropout rate, and the batch size.  

In the BO, the objective function $g: S \to \mathbb{R}$ maps a set of hyper-parameters to a scalar value, representing the performance of the model with those hyper-parameters. In our case, the objective function can be formulated as follows:

\begin{equation}
g(\theta_h^t) = -\text{Accuracy}(\theta_h^t, \theta_1^{*t}, \theta_2^{*t}, X_1, y_1, X_2, y_2)
\tag{11}
\end{equation}

Here, $(X_1, y_1) = \{(X_1^i, y_1^i) \mid \forall i \in \{1, \ldots, N_1\}\}$ and $(X_2, y_2) = \{(X_2^i, y_2^i) \mid \forall i \in \{1, \ldots, N_2\}\}$. The terms $\theta_1^{*t}, \theta_2^{*t}$ denote the optimal parameters of the proposed model obtained through a 2-fold cross-subject validation. Meanwhile, $\theta_h^t \in S$ represents a point within the search space during the $t$th iteration of the BO. In each iteration of the BO, outlined in Algorithm~2, the subsequent hyper-parameter configuration is chosen based on the acquisition function $a(\theta_h^t;B)$. Conventionally, this function is defined as a balance between the Expected Improvement (EI) and the uncertainty in the estimates of the objective function. Its expression is given by:

\begin{equation}
a(\theta_h^t;B) = EI(\theta_h^t;B) - \kappa \, \sigma(\theta_h^t;B)
\tag{12}
\end{equation}

In this context, $\sigma(\theta_h^t;B)$ represents the standard deviation of the objective function estimates at $\theta_h^t$, and $\kappa$ serves as a trade-off parameter regulating the equilibrium between exploration and exploitation.  

Upon the completion of the BO process, the optimal hyper-parameters $\theta_{h_{cv}}^*$ for the $cv$th fold are determined to yield the highest value of the objective function:

\begin{equation}
\theta_{h_{cv}}^* = \arg\max_{\theta_h \in S} g(\theta_h)
\tag{13}
\end{equation}

The evaluation methodology is encapsulated in Algorithm~3. As depicted in this algorithm, the assessment procedure involves the scrutiny of the model's performance. This is achieved through the exploitation of carefully calibrated hyper-parameters, as obtained from Algorithm~1 and Algorithm~2, represented as $\theta_{h_{cv}}^*$. Ultimately, the model's average accuracy is computed through a (10-2)-fold cross-subject validation, as a derivative of this algorithm.  

In the subsequent analysis, we revisit the proposed method, applying a novel formulation that incorporates data augmentation techniques. This approach is predicated on the assumption that the enhancement of our dataset through augmentation can improve the performance and robustness of the method.

\begin{algorithm}[t]
\caption{Bayesian Optimization (BO) Algorithm for Selecting the Next Hyper-Parameter Configuration}
\label{alg:bo-hyper}
\begin{algorithmic}[1]
\Require Dataset $D$, CV folds $\{D_{cv}\}_{cv=1}^{10}$, BO iterations $T=100$
\Ensure Tuned hyper-parameter $\theta_{h^{\ast}_{cv}}$ for each outer fold $cv$

\For{$cv = 1,2,\dots,10$} 
  \For{$t = 1,2,\dots,100$} 
    \State $\displaystyle B_t \gets \{(\theta_h^{1}, g(\theta_h^{1})), (\theta_h^{2}, g(\theta_h^{2})), \dots, (\theta_h^{t}, g(\theta_h^{t}))\}$ 
    \State Train a Gaussian Process (GP) surrogate model on $B_t$ to approximate $g(\theta_h)$
    \State Define acquisition function 
      $\displaystyle a(\theta_h^{t}; B_t) = EI(\theta_h^{t}; B_t) - \kappa \,\sigma(\theta_h^{t}; B_t)$
    \State $\displaystyle \theta_h^{t+1} \gets \arg\max_{\theta_h^{t}} a(\theta_h^{t}; B_t)$ 
    \State $\displaystyle B_{t+1} \gets B_t \cup \{(\theta_h^{t+1}, g(\theta_h^{t+1}))\}$ 
  \EndFor
  \State $\displaystyle \theta_{h^{\ast}_{cv}} \gets \arg\max_{\theta_h \in S} g(\theta_h)$ 
\EndFor
\end{algorithmic}
\end{algorithm}

\begin{algorithm}[h]
\caption{Evaluation Procedure Algorithm Without Data Augmentation}
\label{alg:evaluation}
\begin{algorithmic}[1]
\For{each iteration, denoted by: $cv = 1,2,\dots,10$ do:}
  \State Assign the test set as $D_{cv}$ and the training set as $D_{\text{train}} = D \setminus D_{cv}$
  \State Train the model using $D_{\text{train}}$, and evaluate it on $D_{cv}$
  \State $(X_{\text{train}}^{i}, y_{\text{train}}^{i}) \in D_{\text{train}}, \, \forall i \in \{1, \dots, N_{\text{train}}\}$
  \State $\theta_{cv}^{\ast} = \arg\min_{\theta_{cv}} \sum_{i=1}^{N_{\text{train}}} \sum_{k=1}^{2} -\log \big( p(l_{k} \mid f_{k}(X_{\text{train}}^{i}; \theta_{h,cv}^{\ast}, \theta_{cv})) \big) \cdot y_{\text{train}}^{i}$, Model Tuned Parameters For Outer Fold $cv$
  \State Where, $(X_{cv}^{i}, y_{cv}^{i}) \in D_{cv}, \, \forall i \in \{1, \dots, N_{cv}\}$
  \State $\text{AVG\_ACC} = \sum_{cv=1}^{10} \Big( \frac{1}{N_{cv}} \times \delta \big( \arg\max_{k} p(l_k \mid f_k(X_{cv}^{i}; \theta_{h,cv}^{\ast}, \theta_{cv}^{\ast})) = y_{cv}^{i} \big) \Big)$, Final Model Accuracy
\EndFor
\end{algorithmic}
\end{algorithm}

\subsection{Data Augmentation}\label{subsec2}

DL architecture necessitates extensive datasets for effective training. However, in some research fields such as one presented in this paper, data collection is a time-consuming and laborious task which needs special equipment and facility to capture data in a controlled manner. To address this challenge, a widely adopted technique in the field of DL is DA. It is particularly useful for improving the performance of classification tasks and involves generating new training samples by applying various transformations to the original data, thereby increasing the size and diversity of the dataset. This process helps to mitigate overfitting and enhance the generalization of the model, especially when dealing with limited or imbalanced data \cite{Shorten2019}. In the context of EEG signals, DA has emerged as a promising approach to address the challenges associated with the inherent variability and non-stationarity of these signals, which can hinder the accurate detection of neurological disorders such as ADHD.

Several DA techniques have been proposed for EEG signals, including time-domain, frequency-domain, and spatial transformations \cite{Zhang2020}. One common approach is the addition of Gaussian noise which involves perturbing the raw EEG signals with random noise sampled from a Gaussian distribution. This method has been shown to improve the robustness of DL models by forcing them to learn more discriminative features, rather than relying on the specific characteristics of the training data \cite{Roy2019}. Additional strategies like time-warping, amplitude adjustment, and channel-specific modifications have been investigated to boost the effectiveness of DL discriminators within the scope of EEG-driven ADHD identification \cite{Alhussein2020}.

Our research concentrates on DA application through AGN for raw EEG data in the DL categorization of ADHD. By incorporating this approach, we aim to improve the model's ability to generalize across different EEG recordings and ultimately contribute to the development of more accurate and reliable diagnostic tools for ADHD. Further investigation of other DA techniques and their potential synergistic effects with Gaussian noise may also provide valuable insights into the optimal strategies for enhancing the performance of DL classifiers in this domain.

To generate an augmented sample, we consider $X^i, X_{\text{noise}} \in \mathbb{R}^{E \cdot T}$, where $i$ denotes the $i$th trial of the dataset. Here, $E$ represents the number of electrodes, and $T$ represents the number of time-steps. It is presumed that $X_{j}^{i} \sim \mathcal{N}(0, \sigma^2)$ originates from a Gaussian distribution characterized by a mean of zero ($\mu=0$) and standard deviation ($\sigma$). The magnification level of the AGN is represented by $m$, which can take values from the set $\{1,2,3\}$. The standard deviation $\sigma$ can take values from the set $\{0.1,0.01,0.001\}$. With these parameters, the augmented sample is obtained using the following equation:

\begin{equation}
\tilde{X}^i = X^i + m \times X_{\text{noise}}
\tag{14}
\end{equation}

This equation represents the addition of the original sample $X^i$ and the Gaussian noise $X_{\text{noise}}$, scaled by the magnification level $m$, to create the augmented sample $\tilde{X}^i$. In this study, various combinations of $m$ and $\sigma$ were employed to augment the original data. These combinations can be represented as the union of Cartesian products of different subsets of $M$ and $S$.  

\begin{itemize}
    \item Cartesian products of single elements from $M$ and $S$:  
    \[
    S: (M \times S) = \{(m,\sigma) \mid m \in M, \, \sigma \in S\}
    \]

    \item Cartesian products of pairs of elements from $M$ and $S$:  
    \[
    S: (M \times S)^2 = \{((m_1,\sigma_1),(m_2,\sigma_2)) \mid m_1,m_2 \in M, \, \sigma_1,\sigma_2 \in S, \, m_1 \neq m_2, \, \sigma_1 \neq \sigma_2\}
    \]
\end{itemize}

The final set of combinations can be represented as the union of these Cartesian products:

\begin{equation}
\text{Combinations} = (M \times S) \cup (M \times S)^2 = \{C_1, C_2, \ldots, C_{18}\}
\tag{15}
\end{equation}

Data augmentation (DA) is utilized only during the training phase. After tuning the hyper-parameters of the model during each cycle of the 10-fold cross-subject validation approach, the training data is augmented using various combinations of $m$ and $\sigma$. The model is then trained using the augmented dataset and assessed against the test dataset.  

The assessment procedure utilizing DA is outlined in Algorithm~4. As delineated in this algorithm, the evaluation encompasses the appraisal of the model's performance. This is conducted by leveraging the meticulously tuned hyper-parameters, as derived from Algorithm~1 and Algorithm~2, denoted as $\theta_{h_{cv}}^*$. Furthermore, parameters denoted as $\theta_{cv}^{c*}$ and $\text{Ave\_Acc}_c$ denote optimal parameters of the model tailored for each fold and each respective $C_c$, and the model's average accuracy tailored for each $C_c$, respectively.

\begin{algorithm}[t]
\caption{Training and Testing Procedure With Data Augmentation}
\label{alg:train-test-aug}
\begin{algorithmic}[1]
\For{each iteration, denoted by: $cv = 1,2,\dots,10$ do:}
  \State Assign the test set as $D_{cv}$ and the training set as $D_{\text{train}} = D \setminus D_{cv}$
  \For{each combination of $m,\sigma$ in Combinations set denoted by $C_c, \forall c \in \{1,\dots,18\}$ do:}
    \If{$C_c \in (M \times S)$}
      \State Let $C_c = (m,\sigma)$
      \State Sample $X_{\text{noise}} \sim \mathcal{N}(0,\sigma)$
      \State $X_{\text{train}}^{\prime i} = X_{\text{train}}^{i} + m \times X_{\text{noise}}$ (Augmented Sample)
      \State $D_{\text{train}} = D_{\text{train}} \cup X_{\text{train}}^{\prime i}$ (Updating the Train set)
    \EndIf
    \If{$C_c \in (M \times S)^2$}
      \State $C_c = ((m_{1},\sigma_{1}), (m_{2},\sigma_{2}))$
      \State Sample $X_{\text{noise}}^{(1)} \sim \mathcal{N}(0,\sigma_{1}), \; X_{\text{noise}}^{(2)} \sim \mathcal{N}(0,\sigma_{2})$
      \For{$k = 1,\dots,4$}
        \State $X_{\text{train}}^{\prime i (k)} = X_{\text{train}}^{i} + m_{\lfloor (k+1)/2 \rfloor} \times X_{\text{noise}}^{(\lfloor k/2 \rfloor+1)}$ (Augmented Sample)
        \State $D_{\text{train}} = D_{\text{train}} \cup X_{\text{train}}^{\prime i (k)}$ (Updating the Train set)
      \EndFor
    \EndIf
    \State Train the model using $D_{\text{train}}$, and evaluate it on $D_{cv}$
    \State Where, $(X_{\text{train}}^{i}, y_{\text{train}}^{i}) \in D_{\text{train}}, \forall i \in \{1,\dots,N_{\text{train}}\}$
    \State $\theta_{cv}^{c\ast} = \arg\min_{\theta_{cv}} \sum_{i=1}^{N_{\text{train}}} \sum_{k=1}^{2} -\log \big(p(l_k \mid f_k(X_{\text{train}}^{i}; \theta_{h,cv}^{\ast}, \theta_{cv}^{c}))\big) \cdot y_{\text{train}}^{i}$, Model Tuned Parameters For Outer Fold $cv$ and combination $c$
    \State Where, $(X_{cv}^{i}, y_{cv}^{i}) \in D_{cv}, \forall i \in \{1,\dots,N_{cv}\}$
    \State $\text{AVG\_ACC}_{c} = \sum_{cv=1}^{10} \Big( \frac{1}{N_{cv}} \times \delta \big( \arg\max_{k} p(l_k \mid f_k(X_{cv}^{i}; \theta_{h,cv}^{\ast}, \theta_{cv}^{c\ast})) = y_{cv}^{i} \big) \Big)$
  \EndFor
\EndFor
\end{algorithmic}
\end{algorithm}

\section{Results}\label{sec3}

\subsection{Performance Evaluation}\label{subsec2}
In this study, the performance of the model was assessed in terms of its ability to accurately identify both samples and individuals. To achieve this, two evaluation metrics were utilized: Accuracy and the F2-measure. Considering the greater importance of False Negatives (FN) over False Positives (FP) in the context of this research, the F2-measure offers a broader assessment of the model's efficacy in comparison to Accuracy or F1- measure. The definitions of Classification Accuracy and the F2-measure are provided below, with reference to True Positives (TP), FP, True Negatives (TN), and FN.

Accuracy: the ratio of accurately categorized cases (including both affirmative and negative outcomes) relative to the overall number of cases.  

\begin{equation}
\text{Accuracy} = \frac{TP + TN}{TP + FP + TN + FN}
\tag{16}
\end{equation}

F2-Measure: a variant of the F-measure (F1 score) that gives more importance to recall than precision. It represents the harmonic mean of precision and recall, placing greater emphasis on recall.  

\begin{equation}
F_2 = \frac{(1 + \beta^2) \cdot (\text{Precision} \cdot \text{Recall})}{\beta^2 \cdot \text{Precision} + \text{Recall}}
\tag{17}
\end{equation}

where $\beta$ is a weighting factor (in this case, $\beta = 2$ for the F2-measure), and precision and recall are calculated as outlined:  

\begin{equation}
\text{Precision} = \frac{TP}{TP + FP}
\tag{18}
\end{equation}

\begin{equation}
\text{Recall} = \frac{TP}{TP + FN}
\tag{19}
\end{equation}

The Accuracy and F2-measure are determined for both samples and individuals. For each individual, the predicted label is ascertained utilizing Equation~(4).

The performance comparison between the models, ADHDeepNet and EEGNet, under the purview of the proposed methodology sans DA, is delineated in Table 1 and Table 2. Based on the data presented in these tables, it is discernible that ADHDeepNet has demonstrated superior performance over EEGNet in four metrics investigated in this study. The performance metrics including Sample-Accuracy, Subject-Accuracy, Sample-F2-measure, and Subject-F2-measure for ADHDeepNet were recorded as 0.9013, 0.9256, 0.9373, and 0.9559 respectively. These indicators highlight the superior efficiency of ADHDeepNet in contrast to EEGNet.

\begin{table*}[t]
\caption{The performance of ADHDeepNet using (10-2)-fold cross-subject validation}
\label{tab:adhdeepnet-performance}
\centering
\resizebox{\textwidth}{!}{%
\begin{tabular}{@{}l*{10}{c}c@{}}
\toprule
ADHDeepNet & Fold1 & Fold2 & Fold3 & Fold4 & Fold5 & Fold6 & Fold7 & Fold8 & Fold9 & Fold10 & AVG \\
\midrule
Sample Accuracy      & 0.8745 & 0.8862 & 0.9459 & 0.8825 & 0.8828 & 0.8880 & 0.9185 & 0.9639 & 0.8441 & 0.9256 & 0.9013 ± 0.033 \\
Subject Accuracy     & 0.9167 & 0.8462 & 1.0000 & 0.9167 & 0.9167 & 0.9167 & 0.9167 & 1.0000 & 0.9167 & 0.9167 & 0.9256 ± 0.042 \\
Sample F2-measure    & 0.9600 & 0.9086 & 0.9477 & 0.8715 & 0.9321 & 0.9013 & 0.9680 & 0.9802 & 0.9267 & 0.9749 & 0.9373 ± 0.033 \\
Subject F2 measure   & 0.9722 & 0.8889 & 1.0000 & 0.9524 & 0.9524 & 0.9091 & 0.9615 & 1.0000 & 0.9524 & 0.9756 & 0.9559 ± 0.034 \\
\bottomrule
\end{tabular}%
}
\end{table*}

\begin{table*}[t]
\caption{The performance of EEGNet using (10-2)-fold cross-subject validation}
\label{tab:eegnet-performance}
\centering
\resizebox{\textwidth}{!}{%
\begin{tabular}{@{}l*{10}{c}c@{}}
\toprule
EEGNet & Fold1 & Fold2 & Fold3 & Fold4 & Fold5 & Fold6 & Fold7 & Fold8 & Fold9 & Fold10 & AVG \\
\midrule
Sample Accuracy      & 0.7907 & 0.9119 & 0.8969 & 0.7575 & 0.7929 & 0.8422 & 0.8776 & 0.9149 & 0.7911 & 0.8949 & 0.8499 ± 0.056 \\
Subject Accuracy     & 0.8333 & 0.9230 & 0.9166 & 0.8333 & 0.8333 & 0.8333 & 0.8333 & 0.9166 & 0.8333 & 0.9166 & 0.8677 ± 0.041 \\
Sample F2-measure    & 0.9350 & 0.9053 & 0.8808 & 0.6415 & 0.8858 & 0.8757 & 0.9540 & 0.9589 & 0.9041 & 0.9669 & 0.8931 ± 0.087 \\
Subject F2-measure   & 0.9459 & 0.9090 & 0.8620 & 0.7500 & 0.9090 & 0.8888 & 0.9259 & 0.9615 & 0.9090 & 0.9756 & 0.9037 ± 0.056 \\
\bottomrule
\end{tabular}%
}
\end{table*}

\begin{table*}[t]
\caption{The optimal and baseline performance of the models using data augmentation}
\label{tab:da-performance}
\centering
\resizebox{\textwidth}{!}{%
\begin{tabular}{@{}lcccc@{}}
\toprule
Model & Sample Accuracy & Subject Accuracy & Sample F2-measure & Subject F2-measure \\
\midrule
ADHDeepNet (Optimal)   & 0.9777 ± 0.033 & 0.9917 ± 0.024 & 0.9817 ± 0.033 & 0.9952 ± 0.014 \\
ADHDeepNet (Baseline)  & 0.9058 ± 0.037 & 0.9338 ± 0.032 & 0.9330 ± 0.028 & 0.9580 ± 0.030 \\
EEGNet (Optimal)       & 0.9532 ± 0.039 & 0.9669 ± 0.040 & 0.9711 ± 0.023 & 0.9819 ± 0.022 \\
EEGNet (Baseline)      & 0.8888 ± 0.035 & 0.8925 ± 0.038 & 0.9181 ± 0.067 & 0.9195 ± 0.066 \\
\bottomrule
\end{tabular}%
}
\end{table*}

In the context of the Algorithm~4, the ADHDeepNet and EEGNet models underwent training processes involving various combinations of $m$ and $\sigma$, aiming to evaluate the performance of these models under DA conditions. As presented in Table~3, a comprehensive analysis of the models' performance was conducted, revealing the optimal and baseline performance outcomes. Notably, under the influence of DA, the ADHDeepNet model demonstrated performance efficiency exceeding 97\% across all four previously mentioned metrics, including achieving a sensitivity of 100\%, an accuracy of 99.17\%, and an AUC of 99.38\% in diagnosing ADHD/HC subjects. A low error rate of 0.41\% in sample accuracy between the training and validation sets was observed, affirming the effectiveness of the strategies implemented to mitigate overfitting. This significant result suggests a novel benchmark in the detection of ADHD through DL methodologies. Superior efficacy was ascertained in conjunction with DA parameters comprising $m=1$ and $\sigma=0.001$. Conversely, the poorest performance was correlated with the model conditions $m=3$ and $\sigma=0.1$. As depicted in Fig.~5, the comparative performance assessment underlines the differential outcomes of EEGNet and ADHDeepNet, demonstrating their highest and lowest efficiency points.  

\begin{figure}[t]
\centering
\includegraphics[width=1\textwidth]{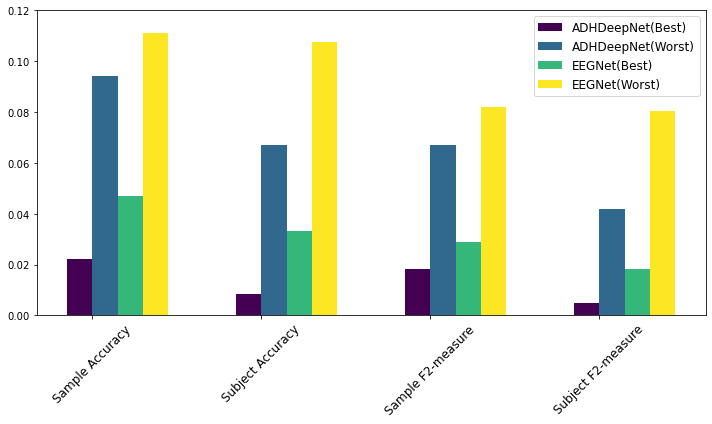}
\caption{The optimal and suboptimal error of the EEGNet and ADHDeepNet using DA}\label{fig5}
\end{figure}

Data augmentation (DA), such as the addition of Gaussian noise to EEG time-series data, can enhance model accuracy through several mechanisms. Firstly, it expands the variety within the training dataset, allowing the model to discern more resilient patterns and mitigating the risk of overfitting. Secondly, DA serves as an implicit regularization technique, akin to $L_1$ and $L_2$ regularization, which prevents overfitting by reinforcing model robustness to variations in the input data. Lastly, this augmentation strategy bolsters the model's capacity for generalization, as exposure to a broader array of data variations during training equips the model to better handle novel, unseen data in the test set. It is hence demonstrable that the proposed method with DA majorly exhibits superior performance compared to scenarios without DA. To investigate the impact of each component detailed in the Model Explanation Section on the effectiveness of the ultimate proposed model, we perform an ablation analysis, which is discussed in the Ablation Study Section.

\subsection{Ablation Study}\label{subsec2}

The proposed model, which was inspired by EEGNet and tailored for detecting ADHD subjects among healthy individuals, retained the initial part of EEGNet and introduced several modifications, including the removal of the last separable Conv2D layer, the addition of an InXception module, the incorporation of a SE block, the reintroduction of the separable Conv2D layer, and the inclusion of a final SE block.  

The InXception module, known for its ability to capture multi-scale features and reduce computational complexity, was added to enhance the model's capacity in identifying intricate patterns within the EEG dataset. The SE blocks, on the other hand, were incorporated to improve the model's performance by adaptively recalibrating channel-wise feature responses, enabling the model to concentrate on the most pertinent features essential for the ADHD classification endeavor.  

The ablation study was performed in two separate stages, where we removed the InXception module and/or the SE blocks. Our findings revealed that the removal of either the InXception module or the SE blocks adversely affected the model's performance. The evaluation metrics employed in this study were F2-measure and Accuracy, which demonstrated the significance of the added modules and blocks in enhancing the model's ability to accurately classify ADHD subjects.  

As demonstrated in Table~4, the results provide evidence for the effectiveness of incorporating the InXception module and SE blocks into the model, as they enhance the model's capability to capture complex patterns and focus on relevant features, ultimately leading to improved classification performance in detecting ADHD subjects among healthy individuals.

\begin{table*}[t]
\caption{The performance of the proposed model in ablation study}
\label{tab:ablation-performance}
\centering
\resizebox{\textwidth}{!}{%
\begin{tabular}{@{}lcccc@{}}
\toprule
Model & Sample Accuracy & Subject Accuracy & Sample F2-measure & Subject F2-measure \\
\midrule
ADHDeepNet                  & 0.9012 ± 0.033 & 0.9256 ± 0.042 & 0.9372 ± 0.033 & 0.9558 ± 0.034 \\
With Incep/Without SENet    & 0.8708 ± 0.038 & 0.8760 ± 0.054 & 0.9099 ± 0.050 & 0.9237 ± 0.054 \\
With SENet/Without Incep    & 0.8789 ± 0.045 & 0.8842 ± 0.041 & 0.9172 ± 0.055 & 0.9201 ± 0.061 \\
EEGNet                      & 0.8499 ± 0.056 & 0.8677 ± 0.041 & 0.8931 ± 0.087 & 0.9037 ± 0.060 \\
\bottomrule
\end{tabular}%
}
\end{table*}

\begin{figure}[t]
\centering
\includegraphics[width=1\textwidth]{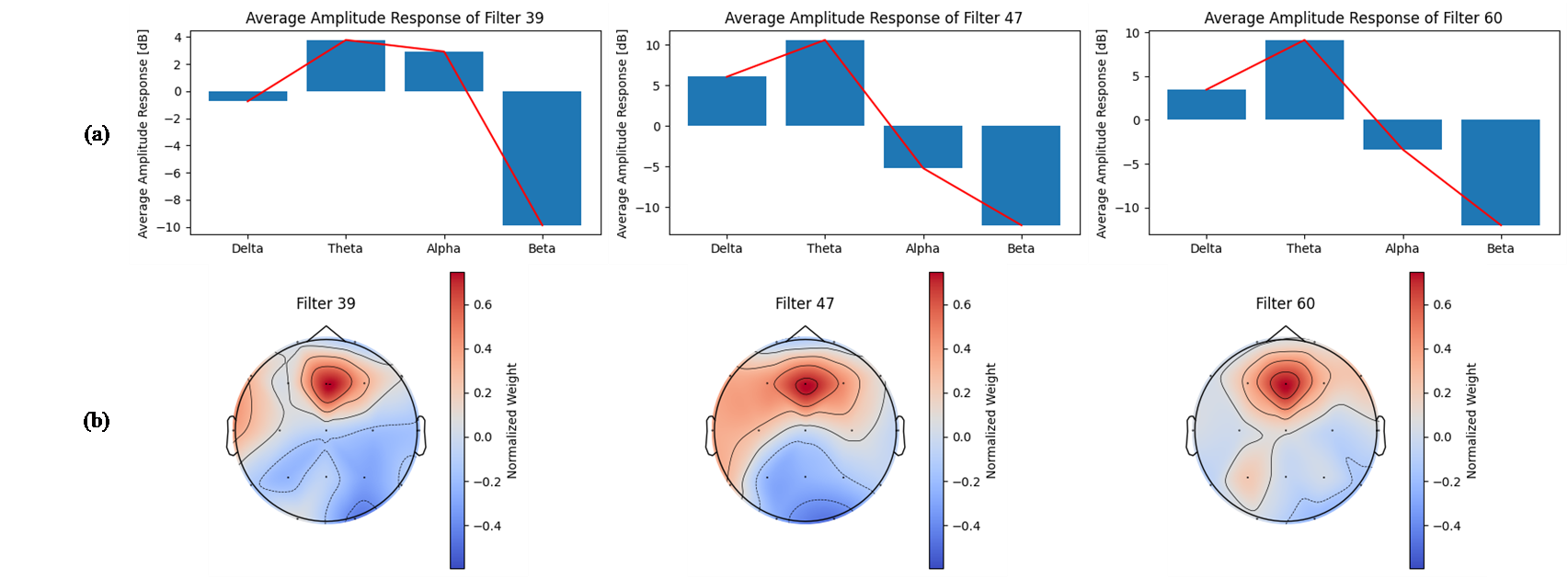}
\caption{(a): Mean amplitude of the frequency response of the filters within the $\delta$, $\theta$, $\alpha$, and $\beta$ bands, (b): The corresponding topographical mapping of brain activity}\label{fig6}
\end{figure}

\section{Discussion}\label{sec3}

Experimental approaches to ADHD detection based on EEG signals can be broadly classified into two main categories: those centered on resting-state tasks and those employing continuous cognitive tasks. For instance, studies in \cite{Chen2019},\cite{Chen2019b} utilize resting-state EEG, whereas other works have investigated continuous-task paradigms. Beyond the task design, these studies differ in their evaluation strategies (cross- vs. within-subject), feature extraction techniques, classifier types, and performance metrics. Table~5 presents a comprehensive comparison of current methodologies for EEG-based ADHD detection.  

Our proposed ADHDeepNet offers several key contributions that address limitations in existing research. First, it processes raw EEG signals directly, thus avoiding intensive manual feature engineering and enabling the network to autonomously capture salient spatiotemporal patterns. Our findings align with broader evidence that temporal–spatial EEG dynamics are meaningful biomarkers, as demonstrated in Alzheimer’s research where reduced flexibility in brain state transitions was linked to disease progression \cite{Razavi2025}. Second, the adaptive attention mechanism (SE blocks) selectively emphasizes relevant frequency bands and spatial regions, increasing the accuracy in identifying ADHD-relevant biomarkers while mitigating noise. Third, we incorporate AGN for data augmentation, expanding the variability of training samples to reduce overfitting and bolster generalization, which is particularly valuable given the challenges of collecting large-scale EEG datasets.  

To avoid data leakage and ensure robustness in real-world scenarios, we implemented a 10-2-fold cross-subject validation. All segments from a single participant are contained within the same fold, preventing sub-signals from the same subject from contaminating both training and testing sets. Specifically, in each outer 10-fold loop, one fold is held out for testing, ensuring robust performance assessment on unseen subjects. Within each outer fold, a 2-fold inner loop is dedicated to hyperparameter tuning, which maintains the independence of the final test fold and mitigates overfitting. This nested cross-validation therefore combines rigorous hyperparameter selection with an unbiased final evaluation.  

Additionally, AGN-based augmentation is applied exclusively to the training set, preventing any overlap of artificial noise parameters with the test data. This design choice preserves an unbiased measure of the model’s real-world applicability and prevents inflated performance metrics that can arise if identical noise characteristics are present in both training and test sets.  

As shown in Table~5, ADHDeepNet surpasses many prior approaches by effectively tackling issues such as reliance on handcrafted features and limiting oneself to within-subject validation schemes. Our method’s combination of raw EEG processing, adaptive attention, spatiotemporal and hierarchical pattern extraction, careful DA with AGN, and nested cross-subject validation yields robust and reliable classification results, highlighting its capability to address shortcomings in earlier methods. Notably, these methodological advancements collectively reduce overfitting and increase sensitivity to subtle EEG differences between ADHD and healthy controls, thus demonstrating a clear performance edge over previously published architectures.

\begin{table*}[t]
\caption{Comparison of cutting-edge methodologies for ADHD detection utilizing EEG signals}
\label{tab:adhd-comparison}
\centering
\resizebox{\textwidth}{!}{%
\begin{tabular}{@{}llllllcc@{}}
\toprule
Study & Evaluation method & cross/within-subject & Feature Extraction & classifier & \multicolumn{2}{c}{Accuracy} & Subjects \\
\cmidrule(lr){6-7}
 & & & & & Sample & Subject & \\
\midrule
\cite{Khare2022} & 10-fold cross-validation & NA & VHERS & ELM & 0.9995 & - & 61 ADHD, 60 HC(*) \\
\cite{Maniruzzaman2023} & 5-fold cross-validation  & NA & TD, Morphological, Non-Linear & GP & 0.9753 & - & 61 ADHD, 60 HC(*) \\
\cite{Khare2023} & 10-fold cross-validation & NA & VMD-HT & EBM & 0.9981 & - & 61 ADHD, 60 HC(*) \\
\cite{Chen2019b} & 8-fold cross-validation  & NA & Power Spectral Density & CNN & 0.9029 & - & 51 ADHD, 57 HC(-) \\
\cite{Chen2019} & 10-fold cross-validation & NA & Connectivity Matrix & CNN & 0.9467 & - & 50 ADHD, 51 HC(-) \\
\cite{Talebi2022}  & 10-fold cross-validation & NA & Non-self Linear Causality Coefficients & ANN & 0.9909 & - & 61 ADHD, 60 HC(*) \\
\cite{Mohammadi2016}  & (70,10,20) Train/Test/Validation & NA & LLE, Fractal Dimension, Entropy & MLP & 0.9365 & - & 31 ADHD, 30 HC(-) \\
\cite{Ekhlasi2023}  & 10-fold cross-validation & cross-subject & Directional Information Transfer & Naïve Bayes & 0.912 & - & 61 ADHD, 60 HC(*) \\
\cite{Allahverdy2016} & (80,20) Train/Test & NA & Lyapunov Exponent, Fractal Dimension & MLP & 0.967 & - & 29 ADHD, 20 HC(-) \\
\cite{Ekhlasi2021}  & 10-fold cross-validation & NA & Directed Phase Transfer Entropy & ANN & 0.892 & - & 61 ADHD, 60 HC(*) \\
\cite{Bakhtyari2022} & 5-fold cross-validation & within-subject & Mel-Spectrogram, DCP & ConvLSTM & 0.9975 & - & 46 ADHD, 45 HC(-) \\
\cite{TaghiBeyglou2022}  & 5-fold cross-validation & cross-subject & Raw EEG Data & CNN+LR & 0.9853 (ADHD) & - & 61 ADHD, 60 HC(*) \\
\cite{Moghaddari2020}  & 5-fold cross-validation & NA & Frequency Band  & CNN & 0.9747 & 0.9848 & 31 ADHD, 30 HC(-) \\
\cite{Tanko2022}  & 10-fold cross-validation & cross-subject & Wavelet and statistical & KNN & 0.9719 & 0.876 & 61 ADHD, 60 HC(*) \\
\textbf{Ours} & \textbf{(10-2)-fold cross-validation} & \textbf{cross-subject} & \textbf{Raw EEG Data} & \textbf{CNN} & \textbf{0.9777} & \textbf{0.9917} & \textbf{61 ADHD, 60 HC} \\
\bottomrule
\end{tabular}%
}
\vspace{2mm}
\begin{minipage}{\textwidth}
\footnotesize
VHERS: Variational Mode Decomposition and Hilbert transform-based EEG rhythm separation,  
TD: Time Domain, VMD-HT: Variational Mode Decomposition-Hilbert Transform,  
ELM: Extreme Learning Machine, GP: Gaussian Process, EBM: Explainable Boosted Machine,  
LLE: Largest Lyapunov Exponent, (*): The same dataset as ours, (-): The abridged or modified version of our dataset
\end{minipage}
\end{table*}

In the process of explaining the ADHDeepNet model, two distinct approaches were employed. The initial technique entailed the visualization of the weights of the primary temporal and depth-wise Conv layer. Subsequently, the t-SNE technique was employed as a means to depict the outputs from different layers within the model. The first approach focused on the direct visualization and interpretation of the Conv layer weights of the model. Understanding the weights of Conv layers is usually a challenging task because each layer is interconnected with many filters from other layers. However, given that ADHDeepNet restricts the connectivity of the Conv layers (via depth-wise and separable Conv), it becomes feasible to interpret the temporal Conv as narrow-band frequency filters and the depth-wise Conv as frequency-specific spatial filters. In this study, we scrutinized the trained weights of the ADHDeepNet within the 8th fold. The initial temporal Conv layer is comprised of 64 kernels, with each one characterized by a shape of $(1 \times 64)$. We interpreted the weights associated with each filter as coefficients of the Conv filter. Consequently, we computed the frequency response for every temporal Conv filter. Subsequently, the mean amplitude of the frequency response of the filters within the delta, theta, alpha, and beta bands was further computed. Moreover, the weights associated with the depth-wise Conv layer corresponding to each of the 64 temporal Conv filters were normalized within the range of $-1$ to $1$. These normalized weights were then mapped onto brain images.  

Let $b_i$ denote the coefficients of the $i$th temporal Conv filter. The frequency response, denoted by $H_i(f)$, was computed using the discrete-time Fourier transform:

\begin{equation}
H_i(f) = \sum_{n=0}^{N-1} b_i[n] \cdot e^{-j 2 \pi f n}
\tag{20}
\end{equation}

where $N$ denotes the count of coefficients, $j$ signifies the imaginary unit, and $f$ is the frequency. Subsequently, the mean amplitude of the frequency response of the filters within the delta ($\delta$), theta ($\theta$), alpha ($\alpha$), and beta ($\beta$) bands was further computed. The average amplitude $A_{\text{avg}, i}$ for the $i$th filter in a specific band is given by:

\begin{equation}
A_{\text{avg}, i}(\text{band}) = \frac{1}{N_{\text{band}}} \sum_{k=1}^{N_{\text{band}}} \left| H_i(f_k) \right|
\tag{21}
\end{equation}

where $N_{\text{band}}$ is the number of frequency points within the specified band. In Fig.~6(a), the average amplitude of the frequency response for a subset of filters across the bands reveals a high theta-to-beta ratio. Concurrently, the corresponding brain mapping images in Fig.~6(b) underscore the activity in the frontal region of the brain, a renowned biomarker for differentiating between ADHD and HC subjects. The findings derived from the ADHDeepNet model align with these observations, suggesting a consistency with established neuroscience principles \cite{Lansbergen2011}.  

To simplify the process, the steps for computing the frequency response and deriving meaningful insights are outlined as follows:  

\begin{itemize}
    \item Each temporal Conv filter’s weights were treated as coefficients of a digital filter.  
    \item The frequency response for each filter was computed using the discrete-time Fourier transform (Equation~20), allowing us to evaluate the filter's response across different frequencies.  
    \item We identified the mean amplitude of the frequency response within the delta, theta, alpha, and beta bands, as shown in Equation~21.  
    \item These band-specific mean amplitudes were visualized to highlight frequency characteristics linked to ADHD (Fig.~6(a)).  
    \item The depth-wise Conv layer’s weights were normalized and mapped onto brain images, providing spatial interpretations of the model’s learned features (Fig.~6(b)).  
\end{itemize}

The second analytical approach involved harnessing the power of t-SNE (t-Distributed Stochastic Neighbor Embedding), a nonlinear technique for reducing dimensionality. This method was strategically employed to distill complex high-dimensional representations within the ADHDeepNet model, focusing on the activation outputs of specific layers. In this study, we visualized and scrutinized the t-SNE-transformed outputs of three distinct layers in ADHDeepNet, providing a granular examination of the model's skill in capturing and discriminating patterns within the ADHD and HC cohorts in the 8th fold.  

The application of t-SNE enabled us to visualize the activation outputs of different layers in the ADHDeepNet model, from the initial layers to the deeper ones. As seen in Fig.~7, the t-SNE-transformed outputs from the initial layers of the model exhibited less discriminatory power between the ADHD and HC groups compared to the outputs of the later layers. This insight implies that the later layers of our ADHDeepNet model are crucial and necessary in distinguishing between the two groups. Therefore, the use of t-SNE not only facilitated the visualization of the model's performance but also highlighted the effectiveness of the deeper layers of the ADHDeepNet in ADHD detection.  

The clinical applicability of the ADHDeepNet model represents a promising avenue for future exploration. The ability to process raw EEG signals with minimal preprocessing and achieve robust, interpretable results positions the model as a potential tool for real-time ADHD diagnosis. Integration into clinical workflows could involve deploying the model in portable EEG devices or software platforms for use in outpatient clinics or schools, enabling rapid, cost-effective screening and early intervention.  

Future steps toward clinical deployment include validating the model on larger, diverse datasets to ensure its generalizability across populations. Additionally, collaborative efforts with clinicians could refine the system for ease of use, ensuring that outputs are interpretable and actionable for healthcare providers. By facilitating early and accurate ADHD detection, ADHDeepNet has the potential to improve treatment outcomes, reduce diagnostic delays, and support personalized interventions tailored to individual needs.

\begin{figure}[t]
\centering
\includegraphics[width=1\textwidth]{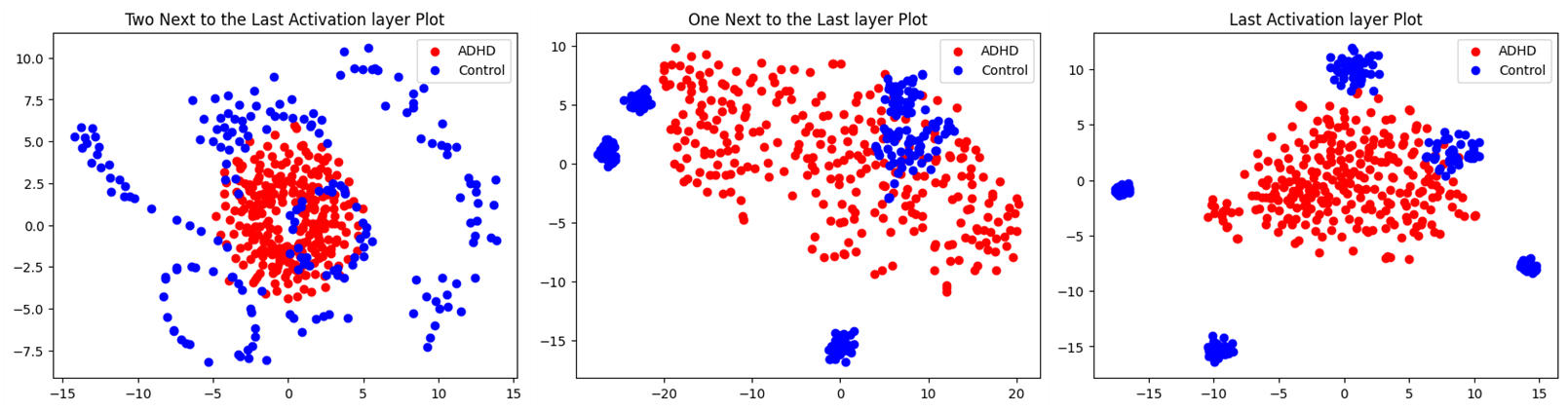}
\caption{Evolution of Discriminatory Power in ADHDeepNet: t-SNE Visualization of Activation Outputs Across Layers for ADHD and HC (8-th Fold)}\label{fig7}
\end{figure}

\section{Conclusion and Future Work}\label{sec3}

This research introduced a DL-based approach, ADHDeepNet, for the diagnosis of ADHD. The model, which is a CNN inspired by the Inception/Xception and attention modules from the SE network, offers a novel approach to ADHD diagnosis. Unlike other machine learning and DL methods, this model does not necessitate pre-processing of EEG data, instead utilizing raw EEG data segmented into four-second intervals. Furthermore, it obviates the necessity for manual derivation of EEG signal traits, thereby positioning itself as a swift and automated solution for ADHD diagnosis using EEG signals.  

To enhance the model's performance and generalizability, AGN was employed for DA. By manipulating the magnification factor and multiple standard deviations $(m,\sigma)$, the original data was augmented during the training phase. The results indicated that varying single and double combinations of different $(m,\sigma)$ significantly impacted the model's performance. It was evident that this method substantially contributed to the enhancement of the model's performance.

Lastly, the model's representation facilitated its explainability and interpretability, particularly in the extraction of discriminative features. This alignment with neuroscience further underscores the potential of the proposed model in the domain of ADHD diagnostics.  

In future research, the utilization of more advanced augmentation strategies, such as generative frameworks based on autoencoders or adversarial models, which have been proposed for biomedical signals such as EEG \cite{Yektaeian2023}, could be explored as a DA strategy. Also, .Additionally, graph-based signal representations, already applied to disorders like obstructive sleep apnea \cite{Vaysi2025}, point toward graph neural networks as a promising avenue for future ADHD EEG research. Outside the EEG domain, sensor-based assessments such as wearable devices combined with dual-task paradigms in stroke survivors \cite{Abdollahi2024} exemplify the broader movement toward quantitative and technology-driven approaches in neurological diagnostics.The efficacy of the proposed methodology could be further evaluated by applying it to additional EEG benchmark datasets, thereby discerning the extent to which this approach can bolster model performance. Moreover, we intend to investigate a potential mathematical correlation between the augmented training data, the original training data, and the test data. This exploration aims to elucidate the underlying mathematical principles of DA, thereby providing a more comprehensive understanding of how this technique can enhance model performance in both within-subject and cross-subject evaluations.

\bibliography{sn-bibliography}

\newpage
\section*{Author Biographies}

\authorbio{Ali Amini}{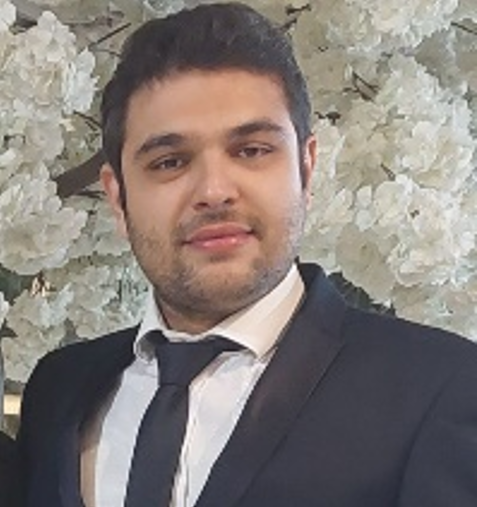}{ALI AMINI received his B.Sc. degree in Electrical Engineering from Shahed University, Tehran, Iran, in 2017, and his M.Sc. degree in Electrical Engineering from Amirkabir University of Technology, Tehran, Iran, in 2021. His master’s thesis focused on using deep learning to enhance surface material classification, which spurred his interest in applying AI methods to complex problems. He is currently a Marie Skłodowska-Curie Ph.D. Fellow at Kaunas University of Technology, Lithuania, researching robust human emotion recognition using hybrid Brain-Computer Interfaces with EEG and EMG signals.}

\authorbio{Mohammad Alijanpour}{me.png}{received the B.S. degree in Electrical engineering from Guilan university, Guilan, Iran, in 2017 and the M.S. degree in Digital Electronic Systems from Amirkabir University of Technology (Tehran Polytechnic), Tehran, Iran, in 2020. He is currently pursuing the Ph.D. degree in Electrical engineering and Computer Science at the University of Central Florida, FL, USA.  
His research interests include deep learning in computer vision, specifically multimodal vision models for video understanding. Mr. Alijanpour’s awards and honors include the ORCGS Doctoral Fellowship from the University of Central Florida, and being awarded as the exceptional talent in his undergraduate studies.
}

\authorbio{Behnam Latifi}{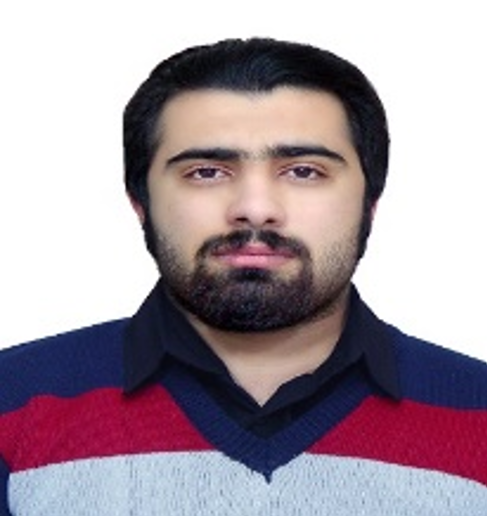}{received the B.Sc. degree in electrical engineering (electronics) from University of Isfahan, Isfahan, Iran, in 2017, and the M.Sc. degree in electrical engineering (digital electronic systems) from Amirkabir University of Technology (Tehran Polytechnic), Tehran, Iran, in 2020. His current research interests include machine learning, deep learning, biomedical signal processing, and image processing.}

\authorbio{ALI MOTIE NASRABADI}{prof.png}{received the B.Sc. degree in electronic engineering and the M.Sc. and Ph.D. degrees in biomedical engineering from the Amirkabir University of Technology, Tehran, Iran, in 1994, 1999, and 2004, respectively. In 2004, he joined the Biomedical Engineering Department, Faculty of Engineering, Shahed University, where he was an Assistant Professor, from 2004 to 2011, an Associate Professor, from 2011 to 2017, and he has been a Full Professor, since 2017. He is currently a Scientific Advisor with the National Brain Mapping Laboratory, University of Tehran, Iran. His current research interests include brain-computer interfaces, biomedical signal processing, machine learning, deep learning, nonlinear time series analysis, and computational neuroscience. He is a Board Member of the Iranian Society for Biomedical Engineering and has served on the scientific committees for several national conferences and review boards of five scientific journals.}

\end{document}